\documentclass[10pt,twocolumn,letterpaper]{article}
\usepackage{iccv}
\usepackage{times}
\usepackage{epsfig}
\usepackage{graphicx}
\usepackage{amsmath}
\usepackage{amssymb}
\usepackage[accsupp]{axessibility}
\usepackage{bm}
\usepackage{multirow}
\usepackage{multicol}
\usepackage{color,soul}
\usepackage[caption=false]{subfig}
\usepackage{microtype}      
\usepackage{appendix}   
\usepackage{algorithm}
\usepackage{algpseudocode}%
\usepackage{booktabs}
\newcommand{\Yt}[1]{\bm{Y}^{#1}}

\newcommand{\Xt}[1]{\bm{X}^{#1}}

\newcommand{\Task}[1]{\mathcal{Z}^{#1}}

\newcommand{\ds}{\textit{MAFPAD}}

\usepackage[pagebackref=true,breaklinks=true,colorlinks,bookmarks=false]{hyperref}

 \iccvfinalcopy 

\def\iccvPaperID{****} 

\ificcvfinal\fi

\begin{document}

\title{Detection and Continual Learning of Novel Face Presentation Attacks}

\author{
Mohammad Rostami$^1$, Leonidas Spinoulas$^1$, Mohamed Hussein$^{1,2}$, Joe Mathai$^1$,  Wael Abd-Almageed$^1$\\
$^1$USC Information Sciences Institute, Los Angeles, CA, 90292 USA\\
$^2$Alexandria University, Alexandria, Egypt\\
{\tt\small \{mrostami,lspinoulas,mehussein,jmathai,wamageed\}@isi.edu}

}

\maketitle
\ificcvfinal\thispagestyle{empty}\fi

\begin{abstract}
  Advances in deep learning, combined with availability of large datasets, have led to impressive improvements in face presentation attack detection research. However, state-of-the-art face antispoofing systems are still vulnerable to novel types of attacks that are never seen during training. Moreover, even if such attacks are correctly detected, these systems lack the ability to adapt to newly encountered attacks. The post-training ability of continually detecting new types of attacks and self-adaptation to identify these attack types, after the initial detection phase, is highly appealing. In this paper, we enable a deep neural network to detect anomalies in the observed input data points as potential new types of attacks by suppressing the confidence-level of the network outside the training samples' distribution. We then use experience replay to update the model to incorporate knowledge about new types of attacks without forgetting the past learned attack types. Experimental results are provided to demonstrate the effectiveness of the proposed method on two benchmark datasets as well as a newly introduced dataset  which exhibits a large variety of attack types.\footnote{Code is available at \url{github.com/mrostami1366}.}
\end{abstract}

\section{Introduction}

Smartphones with facial authentication features have made biometric systems remarkably common in our everyday lives. This is in addition to less prevalent and more traditional, yet critical, applications of biometric systems, such as automatic passport control and access control to high-security facilities.
As services based on biometric recognition technologies gain popularity, presentation attack detection (PAD) is becoming a more crucial requirement for these systems. In parallel, attackers continually attempt to gain unauthorized access by designing new attack types, which makes developing defense mechanisms against presentation attacks more challenging. The goal of a PAD algorithm is to classify whether the presented input to the system is a \emph{bona-fide} presentation (BF) or a presentation attack (PA), so that access is denied for PAs. While recognition systems for all biometric modalities, such as fingerprint and iris, are vulnerable to presentation attacks, the face modality poses a higher risk and extra challenges due to the easy access to high resolution face images of most people, e.g., through social media, and due to the relatively easier fabrication of face PAs. In this paper we exclusively focus on face PAD.

Similar to most sub-fields in computer vision, advances in deep learning, which is inspired by the nervous system~\cite{morgenstern2014properties},  have led to significant face PAD improvements on benchmark datasets using convolutional neural network-based (CNN) end-to-end representation learning and classification~\cite{atoum2017face,jourabloo2018face,liu2018learning,perez2019deep,yang2019face,yu2020searching,wang2020deep, george2021, zitong2020}. Following the standard pipeline of supervised learning for deep learning, a large, labeled training dataset, which consists of known attacks and bona-fide data points, is collected, and then used to train a deep network with a suitable architecture~\cite{krizhevsky2012imagenet,morgenstern2014properties,he2016deep}.

The vulnerability of using the aforementioned pipeline for PAD is that attackers can continually generate new types of PAs that are \emph{unknown} to the system, i.e., absent in the training dataset. Since deep networks suffer from overconfidence in their predictions~\cite{lokhande2020generating}, the system may not be able to identify novel attack types, generated at inference time after the initial training. Even if the unknown attack types can be identified, the standard deep learning pipeline necessitates collecting a sufficient number of samples of new attack types and augmenting the
training dataset. The model then needs to be retrained from scratch (or fine-tuned) on the augmented dataset~\cite{rostami2019sar}. However, collecting labeled data is time consuming, model retraining is computationally inefficient, and both usually involve human intervention~\cite{rostami2018crowdsourcing}. As a result, it is highly desirable to enable biometric recognition systems to identify novel attack types \emph{on-the-fly}, during deployment, and then to autonomously adapt their classifications models for recognizing these new attack types in the future.

We develop an algorithm for continual detection of emerging novel types of face PAs. Our objective is to enable the system to identify novel PAs. The model then is updated to learn new attack types such that it does not forget the past learned attack types. Our idea is based on enabling the network to identify  new attack types as testing samples that are outside the  training distribution samples (OTDS) in an embedding space~\cite{liang2018enhancing,ruff2018deep,yu2019unsupervised,oza2020multiple,george2021}. The base model is then updated to classify these samples as new attacks types in a continual learning (CL) setting, where \textit{catastrophic forgetting}~\cite{french1999catastrophic} is addressed using experience replay~\cite{robins1995catastrophic}. Despite being effective in continual learning (CL) settings, the idea of detecting OTDS has not been explored for PAD.

The main contributions of our work are as follows:
\begin{itemize}
    \item A new formulation of face PAD  as a continual learning problem to equip a PAD system with defense mechanisms that allow learning novel attack types continually.
    \item An algorithm to identify novel attack types as OTDS anomalies by continually screening the input data representations and enable the model to correctly classify them as attacks, in the future, via   experience replay.
    \item A new face anti-spoofing dataset with diverse   attack types to evaluate our algorithm in CL settings.
\end{itemize} 
  
\section{Related Work}
Our work straddles the intersection of two   topics: detection of novel face PAs and continual learning.

\textbf{Novel Presentation-Attack Detection:}
Novel class detection has been studied within several learning settings. The learning setting that we explore is more related to the zero-shot learning (ZSL) formulation~\cite{kolouri2018joint,xie2019attentive,xie2020region,rostami2020using}. ZSL has been studied extensively   but works on  ZSL for PAD have been quite limited. 
Most ZSL works have proposed to identify novel classes using the standard semantic-based idea for describing a sample. In these works, it is assumed that the semantic description of a novel class is accessible \emph{a priori}. Novel classes can then be identified by establishing relationships between known and unknown classes through their semantic descriptions. Note, however, coming up with accurate semantic   descriptions for PAs is challenging.

To relax the need for knowing the  semantic descriptions \emph{a priori},  Shao et al.~\cite{shao2019multi} proposed to learn an embedding space that is discriminative across several source domains to improve generalizability of the PAD model on novel PAs in new domains.
Liu et al.~\cite{liu2019deep} used similarity between an unknown type of attack with known attack types for zero-shot attack detection. 
Both approaches analyze data representations in an embedding space and identify new attack types as unfamiliar data points. Our work follows a similar strategy, where novel attack data points are identified as  OTDS. We then use the collected OTDS to expand the generalizability of the base model on the identified novel attack types.

\textbf{Continual Learning:}
Upon detecting novel attack data points, the base model needs to be continually updated to gain the ability to recognize the newly identified attack types in the future. 
Recent works of CL~\cite{chen2018lifelong} for deep neural networks have mostly focused on tackling \textit{catastrophic forgetting}~\cite{french1999catastrophic}. It occurs when a deep neural network is updated to learn drifts in data distribution in a CL setting which would   lead to underperformance on  the past learned tasks.

Several strategies have been proposed to  mitigate catastrophic forgetting. A group of methods are based on regularizing the deep network weight parameters~\cite{kirkpatrick2017overcoming}. The idea is to identify  the network weights that are important for decent performance on past learned tasks,  consolidate these weights according to their importance,  and learn new tasks using the remaining weights that are unimportant to remember past learned tasks. The main challenge is identifying the important weights and minimize the  negative effects of weight consolidation on the network learning capacity. 
A second group of works are based on the notion of experience replay~\cite{french1999catastrophic}, where the   end-to-end training mechanism of deep networks is changed, rather than the network itself.  The idea is to replay  data points of past learned tasks along with the current task data to update the model through the pseudo-rehearsal process~\cite{robins1995catastrophic}, i.e., retraining the model jointly on the past and the current data. Since  training the network on full datasets is computationally expensive and the storage capacity is limited, only a subset of training data for the past learned tasks should be used. These samples are stored in a memory buffer of fixed maximum capacity~\cite{schaul2015prioritized}. The main challenge is how to select these samples. For example, Schaul et al. select samples that were uncommon and led to maximum learning effects in past tasks~\cite{schaul2015prioritized}. To alleviate the need for a memory buffer, an alternative approach is to use generative models.  The idea is to enable the model to generate pseudo-data points that are similar to the data points of the past tasks and use the pseudo-data points for pseudo-rehearsal~\cite{shin2017continual,rostamicomplementary,rostami2020generative}. We rely on memory-based experience replay to update the PAD model in our work.

\section{Problem Statement}

Consider a PAD task $\Task{0}$ with initial labeled  training data $\mathcal{D}^{0} = \langle \Xt{0}, \Yt{0} \rangle$,   where $  \Xt{0}\in \mathbb{R}^{d \times n_0}$ is the collection of BF data points and a number of fixed known PA instances, and $ \Yt{0} \in \mathbb{R}^{2 \times n_0}$ is the corresponding one-hot binary labels of the PAs and BFs. The training data points  are assumed to be independent and identically distributed (iid) and are drawn from an unknown probability distribution,   $\bm{x}_i^{0}\sim q^{0}(\bm{x})$.  

\begin{figure}[t!]
    \centering
    \includegraphics[width=\linewidth]{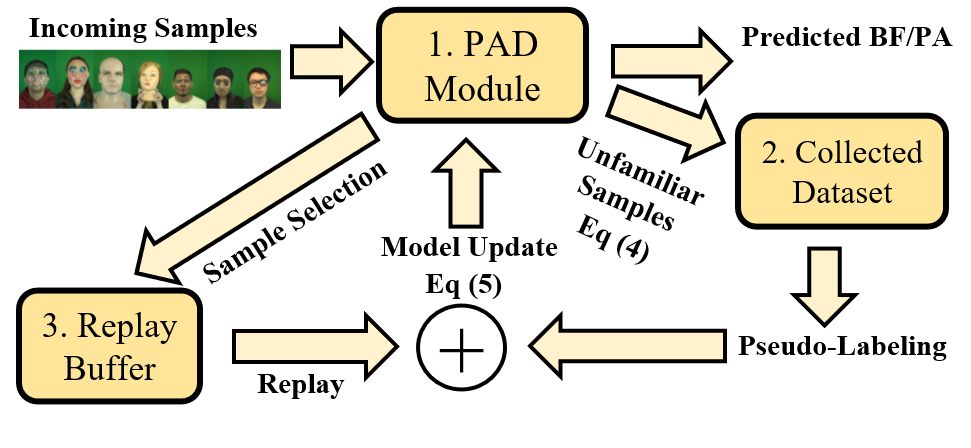}
         \caption{Block-diagram architecture of the proposed continual PAD learning system: 1. PAD module identifies novel attack samples among the input data stream during model execution; 2. The samples from BFs and  PA types are stored to build a dataset; 3. The novel samples and samples stored in the replay buffer are used to update the model through pseudo-rehearsal and the replay buffer.}
         \label{structureFigCVPR2020}
\end{figure}

To solve the initial supervised PAD detection task, we select a parameterized family of functions $f_{\theta}:\mathbb{R}^{ d}\rightarrow \mathbb{R}^{2}$ with learnable  parameters $\theta^{0}$. We then   search for the optimal model  using empirical risk minimization (ERM):
  \begin{equation}
  \small
\begin{split}
 \hat{ \theta}^{0}&=\arg\min_{\theta}\hat{e}_{\theta}= \arg\min_{\theta} \mathbb{E}_{\bm{x}\sim q^{0}(\bm{x})}(\mathcal{L}_d(f^0_{\theta}(\bm{x}),f(\bm{x})))\\&\approx\arg\min_{\theta}\sum_{(\bm{x}_i^{0}, \bm{y}_i^{0}) \in \mathcal{D}^{0}} \mathcal{L}_d(f^0_{\theta}(\bm{x}_i^{0}),\bm{y}_i^{0}),
 \end{split}
\label{eq:CVPR2021ERM}
\end{equation}  
 where $\mathcal{L}_d(\cdot)$  is a proper discrimination loss function such as cross-entropy and $\mathbb{E}(\cdot)$ denotes the probability expectation operator. Upon training the base deep network model on the dataset $\mathcal{D}^{0}$, the PAD system is   fielded for testing. We have depicted this base model as the PAD Module in Figure~\ref{structureFigCVPR2020}.
If $n_0$ is large enough, the selected deep network structure is suitable, and observed data during testing are drawn from the training distribution $q^{0}(\bm{x})$ then the model will generalize well during execution according to theoretical guarantees of PAC-learning framework~\cite{shalev2014understanding}. However, if new types of attacks are introduced after the initial training or any drift in the input distribution occurs, poor model generalization is expected. In other words, the model may fail to identify  new attack types and misclassify them as bona-fide samples. 

To allow for a robust and adaptive PAD system, we extend the standard one-shot training/testing formulation of PAD   to  a continual learning setting~\cite{chen2018lifelong}.
To this end, we consider that after the initial training phase,  PAD tasks  arrive sequentially and we need to address these tasks at inference time. We consider that  the system encounters
sequential   PAD tasks $\{\Task{t}\}_{t=1}^{T_{\text{Max}}}$ in a time sequence $t=1, \ldots, T_{\text{Max}}$ during execution time. Each task is specified by an unlabeled    training dataset $\mathcal{D}^{t} = \langle \Xt{t}\rangle$, $\Xt{t}\in\mathbb{R}^{d\times n_t}$, built from observed input data points over a fixed time period, e.g., a day.  The unlabeled dataset for subsequent tasks may contain new attack types that were not present when the previous tasks were learned, i.e., the tasks may have different distributions $ q^{t}(\bm{x})$. This means that   we need to equip  the PAD module with a mechanism such that the system can identify instances of unknown types of attacks in the dataset $\mathcal{D}^{t}$ at each time-step $t$ and then update the model to learn them (see Figure~\ref{structureFigCVPR2020}). 

If labeled data for BFs and all past learned PAs  is accessible, expanding the model to learn each attack type would be a standard supervised learning problem similar to Eq.~\eqref{eq:CVPR2021ERM}. We just need to augment the dataset $\mathcal{D}^{0}$ with the detected instances of novel PA types and then retrain the base model. However, this would require a memory buffer with unlimited size to store the growing number of observed attack types. Retraining the model continually from scratch can also become computationally expensive and time-consuming.

As a solution, our goal is to update the model by incorporating the new PAs into the system's knowledge by replaying only a subset of training data which are stored in a replay buffer as representative samples (see Figure~\ref{structureFigCVPR2020}).  After updating the model, the system proceeds by learning the subsequent tasks through an iterative procedure. The major challenge of model updating is that, since the past learned attack types may always be encountered, the system must expand its ability to recognize the identified novel attack types such that it maintains the ability to recognize the past learned tasks. This means that the stored samples in the replay buffer need to be  such that they can encode the information required to retain the past tasks knowledge. A high-level block-diagram visualization of our continual learning framework for PAD is provided in Figure~\ref{structureFigCVPR2020}.

\section{Proposed Method}

\begin{figure}[t!]
    \centering
    \includegraphics[]{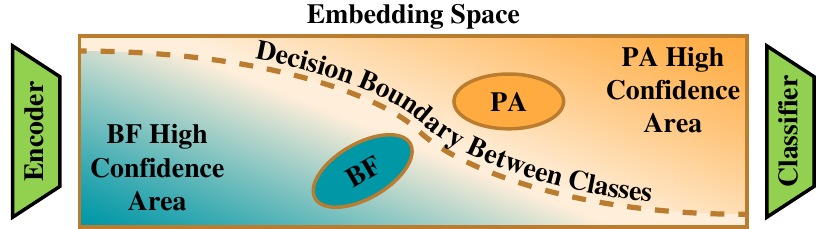}
         \caption{Bimodal data representation  in a discriminative embedding space   after learning a set of PAD types. }
         \label{DataInESFigCVPR2020}
\end{figure}

To solve the challenges of novel attack detection and model updating in a CL setting, we continually screen data representations in a discriminative embedding space which is modeled as the output of a deep encoder. We assume that the deep network can be decomposed into an encoder subnetwork $\phi^t_{\bm{v}}(\cdot):\mathbb{R}^{ d}\rightarrow \mathcal{Z}\subset \mathbb{R}^k$ with learnable parameters $\bm{v}$, e.g., convolutional layers of a CNN, and a classifier subnetwork $h^t_{\bm{w}}(\cdot):\mathbb{R}^{ k}\rightarrow \mathbb{R}^{2}$ with learnable parameters $\bm{w}$, e.g., a sequence of fully connected layers. Here, $\mathcal{Z}$ is  the discriminative embedding space in which the input data points become separable after performing supervised learning. In the case of a deep neural network with good generalization performance,  the embedding space should  be discriminative and the data representation would form a bimodal   distribution~\cite{stan2021unsupervised}, similar to the visualization presented in Figure~\ref{DataInESFigCVPR2020}.

Figure~\ref{DataInESFigCVPR2020} illustrates that the input data distribution is transformed into a bimodal distribution in the embedding space by the encoder subnetwork after learning a PAD task. PAs and BFs each form one mode of this distribution. A decision boundary between these two modes is learned by the classifier subnetwork to classify the input images in the future. The more a data point lies away from the learned decision boundary in the embedding space, the more confident the classifier subnetwork becomes about its prediction. Overconfident area in the embedding space on the BF side of the decision boundary is a major vulnerability of the PAD model (i.e., high-confidence false negatives). If a novel attack is designed such that it lies in this overconfident  region, the model would fail to identify it. Our goal is to make the model robust and stable towards this type of attacks by screening the embedding space, using the intuition above. 
 
 \subsection{Novel Attack Detection}

\begin{figure}[t!]
    \centering
    \includegraphics[width=\linewidth]{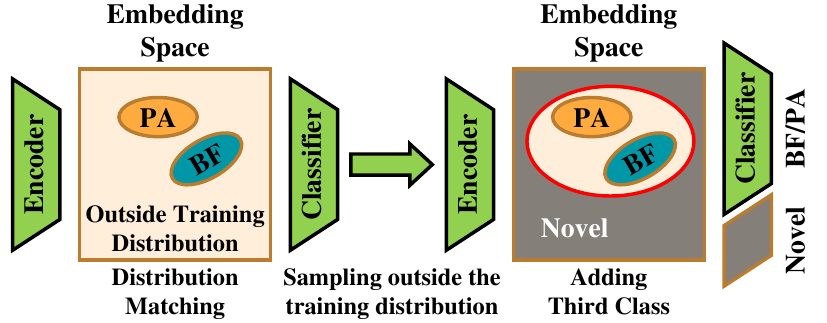}
         \caption{Novel PA detection approach: (left) GMM distribution; (right) adding a third output to the classifier for identifying PAs that may lie in the overconfident regions of the embedding space, visualized as the dark gray region.
         }
         \label{negsamadding}
\end{figure}

To tackle the vulnerability of PAD systems in the overconfident regions, we need to suppress the confidence of the model in those regions. To this end, we fit a parametric distribution to model the learned bimodal distribution in the embedding space. Our idea is based on expanding the base classifier subnetwork and classify the  data points into three classes, namely BFs, PAs, and OTDS (see Figure~\ref{negsamadding}). The intuition behind this idea is that novel PA instances are expected to be different from the training data in the embedding space. This means that we can identify them if the input lies outside the components of the bimodal distribution $\phi(q(\cdot))$ fitted on the embedding. Hence, if we can generate samples that lie outside this distribution, i.e., intuitively the gray region in Figure~\ref{negsamadding}, we can augment the samples from this region with the training data and retrain the classifier subnetwork. As a result, the system will be capable of identifying OTDS data points during execution.

To implement the above rationale, we need to estimate the distribution 
$p(\cdot) = \phi(q^{t}(\cdot))$ before moving forward to start learning at $t+1$.
The empirical version of the learned training distribution at time-step $t$ $\hat{p}^{t}(\cdot)$
is encoded  by the training data representations in the embedding space, $\{(\phi^t_{\bm{v}}(\bm{x}_i^{t}),\bm{y}_i^{t})\}_{i=1}^{N}$ \footnote{We have used a slight abuse of notation. We have assumed that $(\bm{x}_i^{t},\bm{y}_i^{t})$ denotes all the samples that are accessible for training at time $t$. As we will see, these labeled samples consist of novel attack types that have been detected at time $t$, combined with the samples that are selected and stored in the  replay buffer from the previous model update at time $t-1$.} 
Inspired by prototypical networks~\cite{snell2017prototypical},
we   model   $\hat{p}^{t}(\bm{z})$ as a Gaussian mixture model (GMM):
\begin{equation}
\small
\hat{p}^{t}(\bm{z})=\sum_{j=1}^2 \alpha_j^{t} \hat{p}_i^{t}(\bm{z}|j)=\sum_{j=1}^2 \alpha^t_j
\mathcal{N}_j^{t}(\bm{z}|\bm{\mu}_j^{t},\bm{\Sigma}_j^{t}),
\end{equation}  
where $\alpha_j^{t}$ denotes weights for each data modal, i.e., prior probability for BFs and PAs, $\hat{p}_i^{t}(\bm{z}|j)$ is the empirical class conditional probability distribution, and $\bm{\mu}_j^{t}$, $\bm{\Sigma}_j^{t}$ denote the mean and co-variance for each component, respectively. Estimating the GMM parameters is usually performed through expectation maximization (EM)~\cite{moon1996expectation}, which can be a computationally expensive procedure. However, since we have access to labels of data points, we can decouple the GMM components and compute the GMM parameters for each component independently, using MAP estimates. Consider $\bm{S}_j^{t}$ to be the support set for BFs ($j=0$) and PAs ($j=1$) in the training dataset, i.e., $\bm{S}_j^{t}=\{\bm{x}_i^t\in \bm{X}^{t}| \arg\max_c \hat{p}_i^{t}(\bm{x}_i^t|c)=j\}$. Then, we can simply estimate  the GMM parameters as:  
\begin{equation}
\small
\begin{split}
&\hat{\alpha}_j^{t} = \frac{|\bm{S}_j^{t}|}{N}, \qquad \hat{\bm{\mu}}_j^{t} = \frac{1}{|\bm{S}_j^{t}|}\sum_{\bm{x}_i^t\in \bm{S}_j^{t}}\phi^t_v(\bm{x}_i^t), \\& \hat{\bm{\Sigma}}_j^{t} =\frac{1}{|\bm{S}_j^{t}|}\sum_{\bm{x}_i^t\in \bm{S}_j^{t}}\big(\phi^t_v(\bm{x}_i^t)-\hat{\bm{\mu}}_j^{t}\big)^\top\big(\phi^t_v(\bm{x}_i^t)-\hat{\bm{\mu}}_j^{t}\big).
\end{split}
\label{eq:MAPest}
\end{equation}

We rely on the prototypical distributional estimate to generate samples that are outside the training distribution.   We draw random samples from the GMM distribution such that the samples lie in the overconfident region (See Figure~\ref{negsamadding}). For this purpose, we draw random samples from the standard multidimensional Gaussian distribution $\bm{u}\sim \mathcal{N}(\bm{0},\bm{I})$ and then generate samples according to the transformation $\hat{\bm{\mu}}_j+2\hat{\bm{\Sigma}}_j^{\frac{1}{2}} \bm{u}, j = 1, 2$. It is easy to check that these samples are distributed according to the Gaussian distribution of  the $j^{th}$ GMM component. Since we have drawn them to be twice the root of the covariance matrix away from the mean, it is more likely that they lie outside the $j^{th}$ data cluster, as presented in Figure~\ref{negsamadding}.  We use this sampling strategy to generate data points from the overconfident region to expand our model.

Consider that we generate $m$ samples for the $j^{th}$ component as $\bm{X}^j\in\mathbb{R}^{d\times m}$. We fix a probability threshold $\tau < 1$ for the GMM component and then build a pseudo-dataset: 
 \begin{equation}
 \small
\begin{split}
&\bm{X}_{Neg}=[\bm{X}^1_{Neg},\bm{X}^2_{Neg}], \bm{Y}_{Neg}=[\bm{Y}^1_{Neg},\bm{Y}^2_{Neg}] \\& \bm{X}^j_{Neg} = [\bm{x}^j_{1},\ldots \bm{x}^j_{m_j}], \bm{x}^j_k\sim \mathcal{N}(\mu_j,\Sigma_j), j = 1,2\\&
\hat{p}(y=j|\bm{x}^j_k)\le \tau, j =1,2, \bm{y}^j_k = [0, 0, 1]^\top.
\end{split}
\label{eq:pseudodataset}
\end{equation}  
In Eq.~\eqref{eq:pseudodataset}, by using the membership probability, predicted by the GMM,  we ensure to exclude all the generated samples that are close to the means of the GMM components as samples that are inside the distribution. We then build the augmented dataset  $\mathcal{D}^{t}_{aug}=(\bm{X}_{aug}=[\bm{X}_{Neg},\bm{X}^t],\bm{Y}_{aug}=[\bm{Y}_{Neg},\bm{Y}^t])$ for training the ternary classification and then retrain the expanded classifier subnetwork. Note that $(\bm{X}^t,\bm{Y}^t)$ denote the original binary classification   dataset of BFs/PAs on which the  network was initially trained.

As a result of the above process, when the system proceeds to time-step $t+1$ and samples of the dataset $\mathcal{D}^{t+1}$ are encountered during the model execution, the system is able to identify OTDS samples at the third output of the classifier. Let $\mathcal{D}^{t}_{nov}$ denote the OTDS samples  in the dataset $\mathcal{D}^{t}$. We can consider them to be in the attack class. If we retrain the model on the concatenated dataset $\mathcal{D}^{0}\cup\mathcal{D}^{1}_{nov}\cup\ldots\mathcal{D}^{t}_{nov}$, the model would generalize well on the novel attack types. However, this requires storing all observed samples. In the following section, we describe a more efficient approach.

\subsection{Experience Replay for Continual Learning}

To update the model after forming $\mathcal{D}^{t}_{nov}$ at time $t$, we perform experience replay~\cite{robins1995catastrophic} by relying on a replay memory buffer that stores a subset of the observed data after learning each task and before starting learning a subsequent task. Let $\mathcal{D}^{t}_{buf}$ denote the data points stored in the memory buffer (see Figure~\ref{structureFigCVPR2020}). At each batch of optimization, we include samples from both $\mathcal{D}^{t}_{nov}$ and $\mathcal{D}^{t}_{buf}$  in the data batch to update the model. As a result, the model learns to identify novel attacks while retaining the learned knowledge about the past tasks. The only remaining challenge in our framework is a strategy for selecting the samples to be stored in the buffer. 

A simple selection strategy is to randomly select $BD$ samples for each of the BF and the PA classes to store in the memory buffer.
Multiple strategies have been used in the CL literature to improve upon this baseline sampling strategy, including, mean of features (MoF)~\cite{rebuffi2017icarl}, ring buffer~\cite{lopez2017gradient}, and reservoir sampling~\cite{riemer2018learning}. Since we learn the prototypical distribution as a GMM, we can also rely on a strategy similar to MoF. After training the model in the binary classification setting and fitting the GMM, we can compute the distance of all BFs and PAs from their corresponding Gaussian component's mean as $d^t_{j,k}=\|\mu_j^t-\bm{x}^t_k\|_2^2 \mbox{ } \forall \bm{x}^t_k \mbox{ s.t. } \hat{\bm{y}}_k^t=j$.
We sort these distances, for each class separately, and given the per-class memory budget $BD$, we store the samples that are closest to the cluster means. Note that as opposed to a normal CL setting, the labels are predicted for novel PAs in our setting, for $t\ge 1$. Hence, it is more likely that labels for samples close to the means are predicted correctly. However, information about higher moments of the distribution is lost when these samples are used for pseudo-rehearsal. As a result, the model  prediction accuracy may reduce in the area close to the boundary of the classes in the future.

Given the samples stored  in the buffer at $t-1$, we   solve the following pseudo-rehearsal problem for model updating:
 \begin{equation}
 \small
\begin{split}
 \hat{ \theta}^{t}&= \arg\min_{\theta}\Big(\sum_i \mathcal{L}_d(f^t_{\theta}(\bm{x}_{i,nov}^{t}),\bm{y}_{i,nov}^{t})+\\&\sum_i \mathcal{L}_d(f^t_{\theta}(\bm{x}_{i,buff}^{t}),\bm{y}_{i,buff}^{t})+\\&\lambda\sum_i \mathcal{L}_r(\phi^t_{\bm{v}}(\bm{x}_{i,buff}^{t}),\phi^{t-1}_{\bm{v}}(\bm{x}_{i,buff}^{t}) )\Big),
 \end{split}
\label{eq:CVPR2021memorybuffer}
\end{equation}    
where $\lambda$ is a trade-off parameter. The first and the second terms in Eq.~\eqref{eq:CVPR2021memorybuffer} are simply the supervised loss terms for the identified novel samples and the samples stored in the memory buffer, respectively. The third term is added for updating the encoder subnetwork, consistently, according to the past experiences. This term enforces the samples in the memory buffer to be mapped to the proximity of the same location in the embedding space $\phi^{t-1}_{\bm{v}}(\bm{x}_{i,buff}^{t})$ after updating the model to enhance past leaned features. This term can be thought of as a regularization term to mitigate catastrophic forgetting further in addition to  pseudo-rehearsal. Our algorithm, called Novel presentation Attack detection in Continual Learning (NACL), is described in Algorithm~\ref{CVPRPADalgorithm}.

 \begin{algorithm}[t]
\caption{$\mathrm{NACL}\left (\lambda, BD , ITR \right)$\label{CVPRPADalgorithm}} 
 {\small
\begin{algorithmic}[1]
\State \textbf{Initial Training}: 
\State \hspace{2mm}\textbf{Input:} Base dataset $\mathcal{D}^{0}=(\bm{X}^{0},  \bm{Y}^{0})$,
\State \hspace{4mm}\textbf{Initial Training:}
\State \hspace{4mm} $\hat{ \theta}^0=(\hat{\bm{w}}^0,\hat{\bm{v}}^0) =\arg\min_{\theta}\sum_i \mathcal{L}(f_{\theta}(\bm{x}_i^0),\bm{y}_i^0)$
\State \hspace{4mm}\textbf{Prototypical Distribution Estimation:}
\State \hspace{4mm} Use Eq.~\eqref{eq:MAPest} and estimate $\alpha^0_j, \bm{\mu}^0_j,$ and $\Sigma^0_j$
\State \hspace{4mm} Fill the buffer with $\mathcal{D}^0_{buf}$ given the budget $BD$
\State \textbf{Continual Learning}:
\For{$t = 1,\ldots, T_{Max}$ }
\State \textbf{Model Expansion}
\State \hspace{4mm} Use Eq.~\eqref{eq:pseudodataset} to build the pseudo-dataset and then $\mathcal{D}^t_{aug}$
\State \hspace{4mm} Retrain $h^t_{\bm{w}}(\cdot)$ on $\mathcal{D}^t_{aug}$ for ternary classification
\State \textbf{Novel Attack Detection}
\State \hspace{4mm} Build $\mathcal{D}^t_{nov}$ from $\mathcal{D}^t$ to form $\mathcal{D}^t_{nov} \cup \mathcal{D}^{t-1}_{buf}$
\State \hspace{4mm} Update the network weights using Eq.~\eqref{eq:CVPR2021memorybuffer}
\State   \textbf{Prototypical Distribution Estimation:}
\State \hspace{4mm} Use Eq.~\eqref{eq:MAPest} and update $\alpha^t_j, \bm{\mu}^t_j,$ and $\Sigma^t_j$
\State \hspace{4mm} Fill the buffer with $\mathcal{D}^t_{buf}$ given the budget $BD$
\EndFor
\end{algorithmic}}
\end{algorithm}

\section{PADISI-Face Dataset}
\begin{figure}[!t]
    \centering
    \includegraphics[scale=0.48]{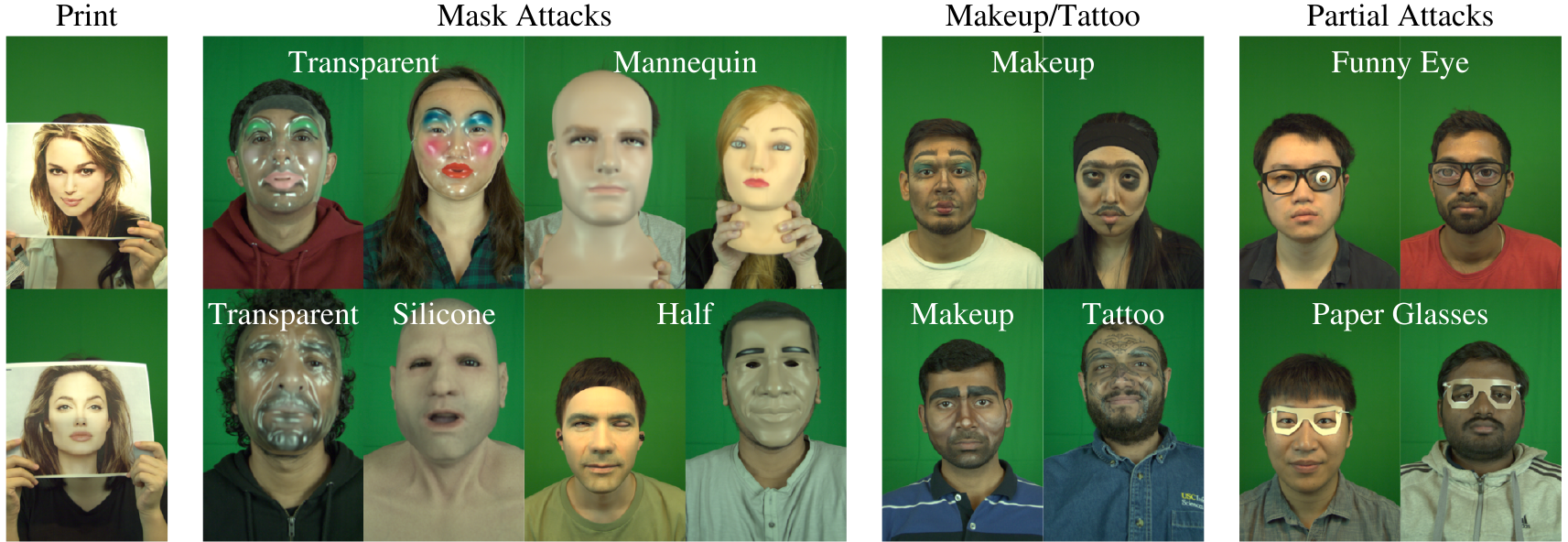}
    \caption{Instances of attacks in the PADISI-Face dataset.}
    \label{fig:attack_samples}
\end{figure}

 To validate our algorithm in a meaningful setup, we need PAD datasets with a diverse set of PAs but such datasets are scarce in the literature. A secondary, yet important, contribution of our work is the introduction of the Face Presentation Attack Detection from Information Sciences Institute (PADISI-Face) dataset  which includes various major face spoofing attack types. To the best of our knowledge, the only other comparable dataset that is accessible at the moment is the recently released HQ-WMCA face anti-spoofing dataset~\cite{idiap_dataset}\footnote{The Wild with Multiple Attacks Database (SiW-M) face anti-spoofing dataset~\cite{liu2019deep} is another existing  dataset with various PA types. SiW-M dataset includes various attack types, similar to PADISI-Face. However, that SiW-M is   temporarily inaccessible.
  Hence, PADISI-Face can  serve as a possible substitute for SiW-M. The PADISI-Face dataset is publicly available  at~\url{https://github.com/ISICV/PADISI_USC_Dataset}.
 }. In PADISI-Face Dataset, each capture consists of a $60$-frame sequence of $1984 \times 1264$ pixel images. The PADISI-Face dataset contains comparable variety of spoofing attacks to HQ-WMCA.  Table~\ref{tab:dataset_statistics}   presents statistics of the collected dataset as well the HQ-WMCA, for comparison.  Figure~\ref{fig:attack_samples} visualizes instances of all the attack types   in the dataset.  For  comprehensive details on the PADISI-Face dataset and its characteristics, please refer to the Appendix.

\begin{table}[!t]
    \centering
    \caption{Summary of statistics of the captured data for the PADISI-Face dataset and comparison to HQ-WMCA~\cite{idiap_dataset}. }
    \label{tab:dataset_statistics}
    
    \resizebox{\linewidth}{!}{
     \setlength\tabcolsep{2pt} 
    \begin{tabular}{c||c|c|c|c|c|c|c} \hline
       Dataset & Participants & \# Captures & \# Frames & Bona-Fide Captures & Attack Captures & Attack Species & Attack Types \\ \hline
      PADISI-Face & $360$ & $2029$ & $121740$ & $1105$ & $924$ & $37$ & $9$ \\ 
        \hline
     HQ-WMCA~\cite{idiap_dataset} & $51$ & $2904$ & $58080$ & $555$ & $2349$ & N/A & $11$    \\ \hline
    \end{tabular}}
\end{table}

\section{Experimental Validation}


 
For our experiments, we adapt suitable benchmark datasets and build incremental PA detection  tasks. Given a dataset with several classes, we assume that the base network is initially trained on a subset of attack types and bona-fide samples. The remaining attack types are observed in a set of sequentially arriving tasks. During each task, new attack types are detected and the model is updated to learn them.

\subsection{Experimental Setup}
 
\textbf{Datasets:} We preform experiments using the    HQ-WMCA~\cite{idiap_dataset} and the new PADISI-Face datasets that are suitable for our learning setting.  The provided unknown attack protocols of these datasets contain only unknown attack types in the testing set and are not suitable for CL setting. As such, we used   the \emph{Grandtest} protocol of HQ-WMCA~\cite{idiap_dataset} to first divide samples into a training and testing set. This protocol contains about $1/3$ of the samples in the test set, with proportional division of each attack type between the training and testing sets, while ensuring that BF samples in the two sets are participant disjoint. For the PADISI-Face dataset, we followed the same division scheme. For both datasets, the CL tasks are constructed using the training set and evaluation is performed on the test set.

\begin{figure*}[t!]
    \centering
    \includegraphics[width=.95\linewidth]{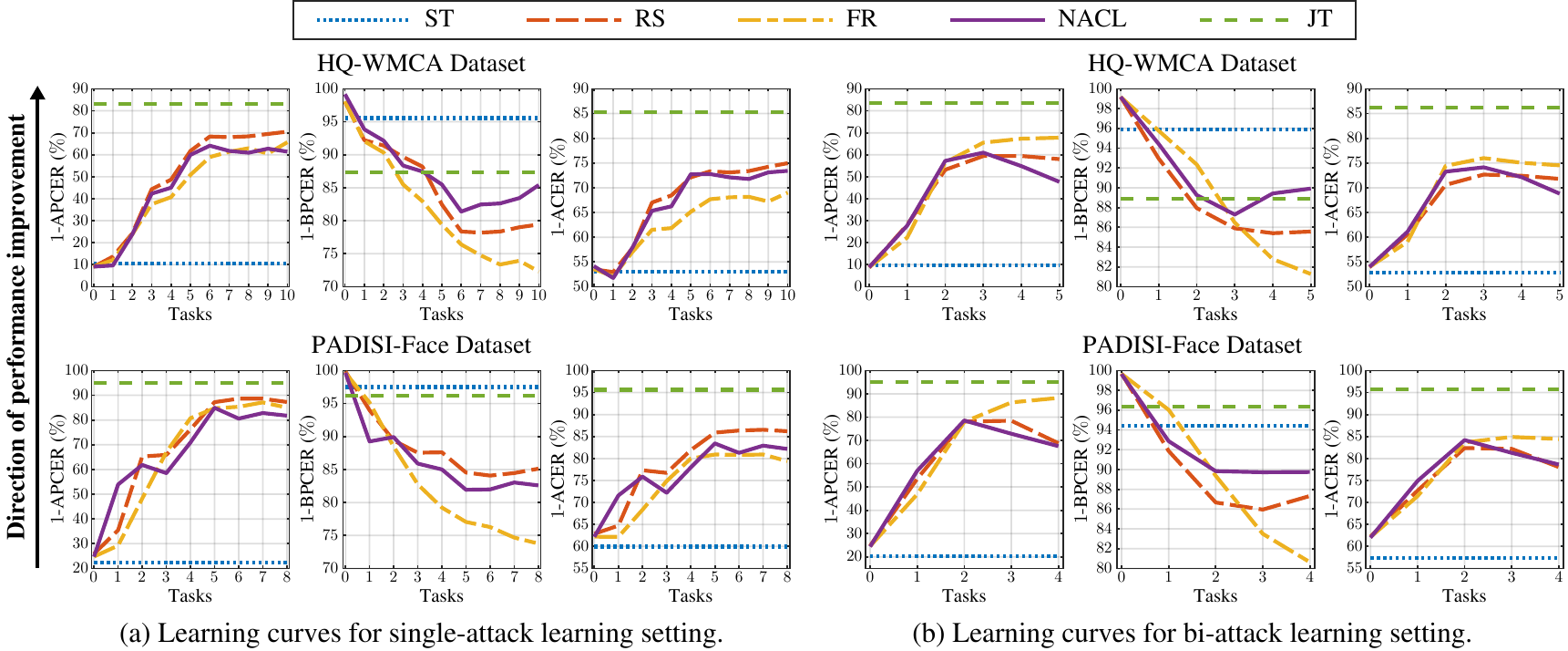}
         \caption{Algorithmic performance for the two experiments (best viewed in color and enlarged on screen).}
         \label{figure:result}
\end{figure*}

\textbf{Baselines for comparison:} Since no prior method in the literature addresses the continual PAD  setting explored in this work, we use three baselines to compare the proposed method with. The presented performance is compared against static training (ST), joint training (JT), and full replay (FR).
In the ST setting, we  report the performance of the base model after initial training without further updating when new attack types in the dataset are encountered. This setting represents performance of existing PAD algorithms when novel PAs are observed and serves as a lower bound. Improvement over this baseline demonstrates relative effectiveness of our approach. In the JT setting, we train the model on the whole labeled training dataset including all attacks types in the initial training. This setting serves an upper-bound which assumes all attack types are known a priori. FR is a variation of Algorithm~\ref{CVPRPADalgorithm} in which we assume that the memory budget is unlimited. As a result, we can save and replay all the stored data points in the buffer. We also report performance of NACL when random sampling (RS) is used to select the buffer samples. In the RS setting, we randomly store selected samples in the memory buffer. Comparison with RS is performed to investigate the effect of using the proposed sampling selection technique. For a fair comparison, we use the same buffer size for both RS and NACL methods. We set the buffer size equal to a fixed size of $100$ samples, filled evenly with BF and PA samples. 

\textbf{Evaluation protocol:} We evaluate the performance of all algorithms using the three standard PAD performance metrics: Attack  Presentation  Classification  Error  Rate (APCER), Bona-Fide Presentation Classification Error Rate (BPCER), and Average Classification Error Rate (ACER). As opposed to the common PAD evaluation setting in which evaluation is performed after training on the full dataset, in one-shot, and only a single number is reported, we generate learning curves to report the PAD performance versus time during execution to encode learning dynamics in our evaluation. In our experiments, we  use the original index-order of the classes  for each dataset, as the order that the attacks are encountered.  At each time-step $t$ , we compute performance of the model on the testing set when the corresponding task is learned and before proceeding to learning the next task.  We report average performance of $10$ randomly initialized runs.

 For   details of the experimental setup, including the network structure, hyper-parameter values,  optimization parameters, and our implementation, please see the Appendix.

\subsection{Results}
Similar to most works in the CL literature, there is a boundary between two subsequent tasks in our formulation. This boundary can be attributed to the instances at which the model is updated after a period of data collection. During each task or period at which the model is not updated, the system may encounter more than one attack types. We consider two sets of experiments for a thorough validation.

First, we consider that the initial training task in our experiments consists of training on bona-fide samples and only the first type of PA, according to the index used in the dataset (see the order of attack types in the Appendix).  Each subsequent task is constructed by introducing one novel attack type. We report the performance of our algorithm and the baselines in Figure~\ref{figure:result}(a). At each time step, we reported the model performance on the full testing split of the datasets.  We have used (1-APCER), (1-BPCER), and (1-ACER) for visualization because learning curves are usually perceived to be increasing functions. Since the testing split is fixed, successful learning is analogous to  rising learning curves.   For a quantitative comparison, we have included the numerical values for the metrics in tabular format in the Appendix.
By inspecting Figure~\ref{figure:result}(a),
  we observe, as expected, that ST is highly vulnerable with respect to novel attack types leading to high values for the APCER and ACER metrics. Note that the high value for BPCER is expected but is not sufficient. This baseline demonstrates the vulnerability of current PAD systems, when novel attacks are encountered, and justifies the necessity of  developing algorithms for PAD in CL settings. When we use the designed novel attack detection mechanism, we can clearly see that performance improves significantly towards the JT upper-bound as more attacks are identified and learned. Performance degradation in terms of BPCER metric is expected due to occurrence of catastrophic forgetting but we see improvements in APCER outweigh this degradation (see ACER plots). Note that RS, FR, and NACL are all equipped with the proposed mechanism and their major difference is in the implementation of the experience replay procedure. We do not see a clear winner between these methods across all metrics but note that NACL and RS offer storing significantly less amount of data in the memory compared to FR (only 100 samples). We also note that in the majority of the time-steps NACL outperforms RS. We conclude that experience replay is an effective approach to address catastrophic forgetting.

  An initially counter-intuitive looking result is that, as opposed to the CL literature, FR does not clearly outperform NACL, despite storing and replaying all samples. However, note that in all RS, FR, and NACL methods, the predicted  labels by the model (not the ground truth) are used in the retraining process. Therefore, FR can be more prone to label pollution, because all samples are stored, leading to performance degradation over time. To verify this intuition, in Table~\ref{labpol}, we provide a comparison of the percentage of polluted labels (stored in the buffer and used for retraining) between the FR, RS, and NACL methods for each learning time-step for the tasks of the PADISI-Face dataset. As observed,  FR indeed faces the challenge of label pollution, leading to performance degradation values similar to RS and NACL. We also observe label pollution is less for NACL at initial time-steps which may explain why after $t=6$ in Figure~\ref{figure:result}(a), learning curves are saturated.  This observation suggests that, as opposed to the normal situation in CL, FR is not necessarily a better option for experience replay even when there is no memory budget limit, due to label pollution.

\begin{table} 
\centering
\label{Tab1}
\caption{Label pollution comparison: percentage of polluted labels per task are reported when learning  PADISI-Face.}
{ \scriptsize
\begin{tabular}{c|cccccccc}
\hline
Task No  & 1 & 2 & 3 & 4 & 5 & 6 & 7 & 8 \\
 \hline
NACL  &   0.0 &  3.2 & 0.5 & 5.7 & 3.8 & 2.0 & 10.5 & 9.8\\
\hline
FR & 0.0&   4.4& 10.0&  13.9& 12.9& 11.4& 10.4 & 9.3 \\
\hline  
RS & 0.0&  15.3 & 5.2 & 8.9 & 9.5 & 9.7& 10.1 &10.6\\
\hline
\end{tabular}
\label{labpol}
}
\end{table}

In the second set of our experiments,
we consider that the initial training task consists of training on bona-fide samples and only the first  PA type, according to the index used in the  datasets. Subsequent tasks are constructed by introducing two novel attack types at each time-step. This setting is closer to a realistic situation. 
We have visualized the learning curves for our algorithm and the baselines in Figure~\ref{figure:result}(b).
Comparing the results with those of Figure~\ref{figure:result}(b), we see that improvements in terms of the ACPER metric are similar. This observation suggests that our algorithm is robust even when multiple attacks are encountered in each time-step.  We also note that performance degradation in terms of the BPCER metric is less than Figure~\ref{figure:result}(a). This observation is expected because the base model has been updated less compared to the single-attack per task scenario. As a result, catastrophic forgetting has been less severe. We conclude that our approach is effective for automatically identifying novel attacks and retraining the model.

\subsection{Analysis and Ablation Studies}

To demonstrate the importance of the ideas used in the NACL algorithm, we   preform ablative experiments.  We considered the   single-attack per task scenario and used the PADISI-Face dataset in these experiments.
We first demonstrate the importance of detecting OTDS samples. Consider that OTDS samples are not detected but the model is updated using a binary prediction baseline. This means that in a CL setting, we always store all the testing samples that are identified as PAs during execution, assuming all to be new attack types, and use them to update the model at each time-step. We refer to this approach as No GMM (NG). In a second experiment, 
we reported performance of the FR setting when real labels (FRR)  are used, i.e., performance in the absence of label pollution. This means that upon identifying the novel attack data points,  rather than using the labels predicted by the model, we use the real labels to update the model. Performance results for these setting are summarized in Table~\ref{tab:continualbatlabl}.
Extremely poor performance of NG, measured in the APCER, demonstrates that detecting OTDS is necessary for PAD in a continual learning setting.  We also observe that, when real labels are used, as expected from the previous discussion, FRR converges to an upper-bound for NACL, close to the visualized JT performance in Figure~\ref{figure:result}(a). This observation suggests a future direction for improving our algorithm is tackling the challenge of label pollution~\cite{natarajan2013learning}. We can also conclude that to mitigate catastrophic forgetting further, a larger buffer size should be used.

\begin{table}[!t]
    \caption{Ablation studies using the   tasks for the PADISI-Face dataset during execution time in the single PA/task scenario.}
    \label{tab:continualbatlabl}
    \centering
    \resizebox{\linewidth}{!}{
    \begin{tabular}{c|cccc|cccc|cccc} \hline
         Task     &  \multicolumn{4}{c|}{APCER ($\%$)}     & \multicolumn{4}{c|}{BPCER ($\%$)}      & \multicolumn{4}{c}{ACER ($\%$)}\\
         \hline 
         No.      &  FRR    & NG    & ED    & DE  &  FRR    & NG    & ED    & DE  &  FRR    & NG    & ED    & DE   \\   \hline
$ 1 $
& $ 66.2 $ & $ 74.6 $ & $ 65.3 $ & $ 62.4 $
& $ 1.2 $ & $ 0.1 $ & $ 4.4 $ & $ 7.1 $
& $ 33.7 $ & $ 37.3 $ & $ 34.8 $ & $ 34.8 $ \\
$ 2 $
& $ 40.2 $ & $ 74.8 $ & $ 37.8 $ & $ 39.7 $
& $ 0.3 $ & $ 0.1 $ & $ 8.8 $ & $ 15.0 $
& $ 20.3 $ & $ 37.5 $ & $ 23.3 $ & $ 27.4 $ \\
$ 3 $
& $ 41.8 $ & $ 74.5 $ & $ 29.8 $ & $ 34.6 $
& $ 0.3 $ & $ 0.0 $ & $ 10.1 $ & $ 18.4 $
& $ 21.0 $ & $ 37.3 $ & $ 19.9 $ & $ 26.5 $ \\
$ 4 $
& $ 31.6 $ & $ 74.6 $ & $ 27.0 $ & $ 24.2 $
& $ 0.3 $ & $ 0.0 $ & $ 11.1 $ & $ 19.4 $
& $ 15.9 $ & $ 37.3 $ & $ 19.1 $ & $ 21.8 $ \\
$ 5 $
& $ 17.7 $ & $ 74.8 $ & $ 23.2 $ & $ 18.0 $
& $ 1.0 $ & $ 0.0 $ & $ 11.1 $ & $ 21.0 $
& $ 9.3 $ & $ 37.4 $ & $ 17.1 $ & $ 19.5 $ \\
$ 6 $
& $ 15.7 $ & $ 75.2 $ & $ 16.4 $ & $ 13.9 $
& $ 0.7 $ & $ 0.0 $ & $ 12.7 $ & $ 21.0 $
& $ 8.2 $ & $ 37.6 $ & $ 14.5 $ & $ 17.5 $ \\
$ 7 $
& $ 14.9 $ & $ 75.0 $ & $ 20.4 $ & $ 13.8 $
& $ 0.6 $ & $ 0.0 $ & $ 13.3 $ & $ 19.9 $
& $ 7.7 $ & $ 37.5 $ & $ 16.9 $ & $ 16.8 $ \\
$ 8 $
& $ 13.0 $ & $ 75.2 $ & $ 24.9 $ & $ 35.3 $
& $ 0.6 $ & $ 0.0 $ & $ 13.6 $ & $ 14.4 $
& $ 6.8 $ & $ 37.6 $ & $ 19.2 $ & $ 24.9 $ \\
 \hline  
    \end{tabular}} 
\end{table}

We also study the effect of the temporal observation-order at which the  PAs are encountered on our algorithm performance. The order we used in our experiments is arbitrary and preset. 
But in practice, the user does not have any control on the temporal order at which the PAs are observed during execution. For this reason, we consider two extreme cases of ordering. We use the pre-update difficulty of PA detection by the model to set a synthetic temporal ordering
on PA types. To this end, we start learning the PA with the class index 1 in the  dataset. After learning the first task, for all time-steps, we  compute the performance of the model on all the remaining PA types. The detection rates for the remaining PAs are a measure of difficulty of detecting (or learning) them by the model. We performed experiments using two  easy to difficult (ED) and difficult to easy (DE) orderings. In the ED scenario, we pick the PA with largest detection rate as the next observed PA.  This PA is the easiest PA for the system to learn among the remaining PAs.  It is the most similar PA to learned PAs from the model's point of view. We continue until all the attacks are observed. In the DE scenario, we pick the PA with the least detection rate.

Results for ED and DE temporal orderings are reported in Table~\ref{tab:continualbatlabl}.
We observe that in both cases, NACL algorithm is able to improve the performance of the model as more PA types are encountered and learned. The final model performance after observing all PAs denotes that learning in the ED scenario is easier for the algorithm. This observation accords with our intuition because learning novel attacks that are less similar to the previously observed attack types is more challenging. We conclude that the particular PA observation ordering influences the performance of our algorithm, but our algorithm is effective   in the worst-case scenario.

Finally, we highlight that our method is stronger is reducing false-negative predictions. In the Appendix, we have demonstrated that by benefiting from manual annotation of the novel samples, i.e., reducing the label pollution, we can  considerably reduce the false-positive predictions.
 
\section{Conclusions}

We study  the problem of PA detection in a continual learning setting. Our proposed approach is based on screening the data representations in an embedding space. We estimate the learned training data distribution in the embedding space using a  GMM distribution. We use this distribution to enable the base model to identify novel attack types as outside training distribution samples.  Experience replay is then used to update the model to tackle catastrophic forgetting. We also collect a new dataset that contains various types of face spoofing attacks. Experiments on two datasets  demonstrate that our method is effective for a continual learning setting. Future research direction includes  tackling label pollution and considering tasks without sharp temporal boundaries.

\section{Acknowledgment}
 This research is based upon work supported by the Office of the Director of National Intelligence (ODNI), Intelligence Advanced Research Projects Activity (IARPA), via IARPA R\&D Contract No. 2017-17020200005. The  views  and  conclusions  contained  herein should not be interpreted as necessarily representing the official policies or endorsements, either expressed or  implied,  of  the  ODNI,  IARPA,  or  the  U.S.  Government. The  U.S.  Government  is  authorized  to  reproduce  and  distribute  reprints  for  Governmental  purposes  not withstanding any copyright annotation thereon.

{\small
\bibliographystyle{ieee_fullname}
\bibliography{ref}

\begin{thebibliography}{10}\itemsep=-1pt

\bibitem{mlfp}
A. {Agarwal}, D. {Yadav}, N. {Kohli}, R. {Singh}, M. {Vatsa}, and A. {Noore}.
\newblock {Face Presentation Attack with Latex Masks in Multispectral Videos}.
\newblock In {\em 2017 IEEE Conference on Computer Vision and Pattern
  Recognition Workshops (CVPRW)}, pages 275--283, 2017.

\bibitem{atoum2017face}
Y. {Atoum}, Y. {Liu}, A. {Jourabloo}, and X. {Liu}.
\newblock Face anti-spoofing using patch and depth-based cnns.
\newblock In {\em 2017 IEEE International Joint Conference on Biometrics
  (IJCB)}, pages 319--328, 2017.

\bibitem{fan}
A. {Bulat} and G. {Tzimiropoulos}.
\newblock {Super-FAN: Integrated Facial Landmark Localization and
  Super-Resolution of Real-World Low Resolution Faces in Arbitrary Poses with
  GANs}.
\newblock In {\em 2018 IEEE/CVF Conference on Computer Vision and Pattern
  Recognition}, pages 109--117, 2018.

\bibitem{chen2018lifelong}
Z. Chen, B. Liu, R. Brachman, P. Stone, and F. Rossi.
\newblock {\em Lifelong Machine Learning}.
\newblock Morgan \& Claypool Publishers, 2nd edition, 2018.

\bibitem{msspoof-2015}
Ivana Chingovska, Nesli Erdogmus, Andr{\'e} Anjos, and S{\'e}bastien Marcel.
\newblock {Face Recognition Systems Under Spoofing Attacks}.
\newblock In Thirimachos Bourlai, editor, {\em Face Recognition Across the
  Imaging Spectrum}, pages 165--194. Springer International Publishing, Cham,
  2016.

\bibitem{I2BVSD}
Tejas~Indulal Dhamecha, Richa Singh, Mayank Vatsa, and Ajay Kumar.
\newblock {Recognizing Disguised Faces: Human and Machine Evaluation}.
\newblock {\em PLOS ONE}, 9(7):1--16, 07 2014.

\bibitem{3dmad}
N. {Erdogmus} and S. {Marcel}.
\newblock {Spoofing in 2D face recognition with 3D masks and anti-spoofing with
  Kinect}.
\newblock In {\em 2013 IEEE Sixth International Conference on Biometrics:
  Theory, Applications and Systems (BTAS)}, pages 1--6, 2013.

\bibitem{french1999catastrophic}
R.~M. French.
\newblock Catastrophic forgetting in connectionist networks.
\newblock {\em Trends in Cognitive Sciences}, 3(4):128--135, 1999.

\bibitem{george2021}
A. {George} and S. {Marcel}.
\newblock Learning one class representations for face presentation attack
  detection using multi-channel convolutional neural networks.
\newblock {\em IEEE Transactions on Information Forensics and Security},
  16:361--375, 2021.

\bibitem{he2020momentum}
K. {He}, H. {Fan}, Y. {Wu}, S. {Xie}, and R. {Girshick}.
\newblock Momentum contrast for unsupervised visual representation learning.
\newblock In {\em 2020 IEEE/CVF Conference on Computer Vision and Pattern
  Recognition (CVPR)}, pages 9726--9735, 2020.

\bibitem{he2016deep}
Kaiming He, Xiangyu Zhang, Shaoqing Ren, and Jian Sun.
\newblock Deep residual learning for image recognition.
\newblock In {\em Proceedings of the IEEE conference on computer vision and
  pattern recognition}, pages 770--778, 2016.

\bibitem{idiap_dataset}
G. {Heusch}, A. {George}, D. {Geissbühler}, Z. {Mostaani}, and S. {Marcel}.
\newblock Deep models and shortwave infrared information to detect face
  presentation attacks.
\newblock {\em IEEE Transactions on Biometrics, Behavior, and Identity
  Science}, 2(4):399--409, 2020.

\bibitem{jourabloo2018face}
A. Jourabloo, Y. Liu, and X. Liu.
\newblock Face de-spoofing: Anti-spoofing via noise modeling.
\newblock In V. Ferrari, M. Hebert, C. Sminchisescu, and Y. Weiss, editors,
  {\em Computer Vision -- ECCV 2018}, pages 297--315, Cham, 2018. Springer
  International Publishing.

\bibitem{kirkpatrick2017overcoming}
J. Kirkpatrick, R. Pascanu, N. Rabinowitz, J. Veness, G. Desjardins, A.~A.
  Rusu, K. Milan, J. Quan, T. Ramalho, A. Grabska-Barwinska, D. Hassabis, C.
  Clopath, D. Kumaran, and R. Hadsell.
\newblock Overcoming catastrophic forgetting in neural networks.
\newblock {\em Proceedings of the National Academy of Sciences},
  114(13):3521--3526, 2017.

\bibitem{kolouri2018joint}
Soheil Kolouri, Mohammad Rostami, Yuri Owechko, and Kyungnam Kim.
\newblock Joint dictionaries for zero-shot learning.
\newblock In {\em Proceedings of the AAAI Conference on Artificial
  Intelligence}, volume~32, 2018.

\bibitem{krizhevsky2012imagenet}
Alex Krizhevsky, Ilya Sutskever, and Geoffrey~E Hinton.
\newblock Imagenet classification with deep convolutional neural networks.
\newblock {\em Advances in neural information processing systems},
  25:1097--1105, 2012.

\bibitem{liang2018enhancing}
S. Liang, Y. Li, and R. Srikant.
\newblock Enhancing the reliability of out-of-distribution image detection in
  neural networks.
\newblock In {\em International Conference on Learning Representations}, 2018.

\bibitem{liu2018learning}
Y. {Liu}, A. {Jourabloo}, and X. {Liu}.
\newblock Learning deep models for face anti-spoofing: Binary or auxiliary
  supervision.
\newblock In {\em 2018 IEEE/CVF Conference on Computer Vision and Pattern
  Recognition}, pages 389--398, 2018.

\bibitem{liu2019deep}
Y. {Liu}, J. {Stehouwer}, A. {Jourabloo}, and X. {Liu}.
\newblock Deep tree learning for zero-shot face anti-spoofing.
\newblock In {\em 2019 IEEE/CVF Conference on Computer Vision and Pattern
  Recognition (CVPR)}, pages 4675--4684, 2019.

\bibitem{lokhande2020generating}
V.~S. {Lokhande}, S. {Tasneeyapant}, A. {Venkatesh}, S.~N. {Ravi}, and V.
  {Singh}.
\newblock Generating accurate pseudo-labels in semi-supervised learning and
  avoiding overconfident predictions via hermite polynomial activations.
\newblock In {\em 2020 IEEE/CVF Conference on Computer Vision and Pattern
  Recognition (CVPR)}, pages 11432--11440, 2020.

\bibitem{lopez2017gradient}
D. Lopez-Paz and M.'A. Ranzato.
\newblock Gradient episodic memory for continual learning.
\newblock In {\em Proceedings of the 31st International Conference on Neural
  Information Processing Systems}, NIPS'17, page 6470–6479, Red Hook, NY,
  USA, 2017. Curran Associates Inc.

\bibitem{moon1996expectation}
T.~K. {Moon}.
\newblock The expectation-maximization algorithm.
\newblock {\em IEEE Signal Processing Magazine}, 13(6):47--60, 1996.

\bibitem{morgenstern2014properties}
Yaniv Morgenstern, Mohammad Rostami, and Dale Purves.
\newblock Properties of artificial networks evolved to contend with natural
  spectra.
\newblock {\em Proceedings of the National Academy of Sciences}, 111(Supplement
  3):10868--10872, 2014.

\bibitem{natarajan2013learning}
Nagarajan Natarajan, Inderjit~S Dhillon, Pradeep Ravikumar, and Ambuj Tewari.
\newblock Learning with noisy labels.
\newblock In {\em NIPS}, volume~26, pages 1196--1204, 2013.

\bibitem{oza2020multiple}
P. Oza, H.~V. Nguyen, and V.~M. Patel.
\newblock Multiple class novelty detection under data distribution shift.
\newblock In A. Vedaldi, H. Bischof, T. Brox, and J.-M. Frahm, editors, {\em
  Computer Vision -- ECCV 2020}, pages 432--449, Cham, 2020. Springer
  International Publishing.

\bibitem{nir_dataset_01}
I. {Pavlidis} and P. {Symosek}.
\newblock {The imaging issue in an automatic face/disguise detection system}.
\newblock In {\em Proceedings IEEE Workshop on Computer Vision Beyond the
  Visible Spectrum: Methods and Applications (Cat. No.PR00640)}, pages 15--24,
  2000.

\bibitem{perez2019deep}
D. {Pérez-Cabo}, D. {Jiménez-Cabello}, A. {Costa-Pazo}, and R.~J.
  {López-Sastre}.
\newblock Deep anomaly detection for generalized face anti-spoofing.
\newblock In {\em 2019 IEEE/CVF Conference on Computer Vision and Pattern
  Recognition Workshops (CVPRW)}, pages 1591--1600, 2019.

\bibitem{GUC-LiFFAD}
R. {Raghavendra}, K.~B. {Raja}, and C. {Busch}.
\newblock {Presentation Attack Detection for Face Recognition Using Light Field
  Camera}.
\newblock {\em IEEE Transactions on Image Processing}, 24(3):1060--1075, 2015.

\bibitem{emspad}
R. {Raghavendra}, K.~B. {Raja}, S. {Venkatesh}, F.~A. {Cheikh}, and C. {Busch}.
\newblock {On the vulnerability of extended Multispectral face recognition
  systems towards presentation attacks}.
\newblock In {\em 2017 IEEE International Conference on Identity, Security and
  Behavior Analysis (ISBA)}, pages 1--8, 2017.

\bibitem{rebuffi2017icarl}
S. {Rebuffi}, A. {Kolesnikov}, G. {Sperl}, and C.~H. {Lampert}.
\newblock {iCaRL}: Incremental classifier and representation learning.
\newblock In {\em 2017 IEEE Conference on Computer Vision and Pattern
  Recognition (CVPR)}, pages 5533--5542, 2017.

\bibitem{riemer2018learning}
M. Riemer, I. Cases, R. Ajemian, M. Liu, I. Rish, Y. Tu, , and G. Tesauro.
\newblock Learning to learn without forgetting by maximizing transfer and
  minimizing interference.
\newblock In {\em International Conference on Learning Representations}, 2019.

\bibitem{robins1995catastrophic}
A. Robins.
\newblock Catastrophic forgetting, rehearsal and pseudorehearsal.
\newblock {\em Connection Science}, 7(2):123--146, 1995.

\bibitem{rostami2018crowdsourcing}
Mohammad Rostami, David Huber, and Tsai-Ching Lu.
\newblock A crowdsourcing triage algorithm for geopolitical event forecasting.
\newblock In {\em Proceedings of the 12th ACM Conference on Recommender
  Systems}, pages 377--381, 2018.

\bibitem{rostami2020using}
Mohammad Rostami, David Isele, and Eric Eaton.
\newblock Using task descriptions in lifelong machine learning for improved
  performance and zero-shot transfer.
\newblock {\em Journal of Artificial Intelligence Research}, 67:673--704, 2020.

\bibitem{rostami2019sar}
Mohammad Rostami, Soheil Kolouri, Eric Eaton, and Kyungnam Kim.
\newblock Sar image classification using few-shot cross-domain transfer
  learning.
\newblock In {\em Proceedings of the IEEE/CVF Conference on Computer Vision and
  Pattern Recognition Workshops}, pages 0--0, 2019.

\bibitem{rostami2020generative}
Mohammad Rostami, Soheil Kolouri, Praveen Pilly, and James McClelland.
\newblock Generative continual concept learning.
\newblock In {\em Proceedings of the AAAI Conference on Artificial
  Intelligence}, volume~34, pages 5545--5552, 2020.

\bibitem{rostamicomplementary}
Mohammad Rostami, Soheil Kolouri, and Praveen~K Pilly.
\newblock Complementary learning for overcoming catastrophic forgetting using
  experience replay.
\newblock In {\em Proceedings of the International Joint Conference on
  Artificial Intelligence}, pages 3339--3345, 2019.

\bibitem{ruff2018deep}
L. Ruff, R. Vandermeulen, N. Goernitz, L. Deecke, S.~A. Siddiqui, A. Binder, E.
  M{\"u}ller, and M. Kloft.
\newblock Deep one-class classification.
\newblock In J. Dy and A. Krause, editors, {\em Proceedings of the 35th
  International Conference on Machine Learning}, volume~80 of {\em Proceedings
  of Machine Learning Research}, pages 4393--4402, Stockholmsmässan, Stockholm
  Sweden, 10--15 Jul 2018. PMLR.

\bibitem{schaul2015prioritized}
T. Schaul, J. Quan, I. Antonoglou, and D. Silver.
\newblock Prioritized experience replay.
\newblock {\em ArXiv}, abs/1511.05952, 2015.

\bibitem{shalev2014understanding}
S. Shalev-Shwartz and S. Ben-David.
\newblock {\em Understanding Machine Learning: From Theory to Algorithms}.
\newblock Cambridge University Press, USA, 2014.

\bibitem{shao2019multi}
R. {Shao}, X. {Lan}, J. {Li}, and P.~C. {Yuen}.
\newblock Multi-adversarial discriminative deep domain generalization for face
  presentation attack detection.
\newblock In {\em 2019 IEEE/CVF Conference on Computer Vision and Pattern
  Recognition (CVPR)}, pages 10015--10023, 2019.

\bibitem{shin2017continual}
H. Shin, J.~K. Lee, J. Kim, and J. Kim.
\newblock Continual learning with deep generative replay.
\newblock In I. Guyon, U.~V. Luxburg, S. Bengio, H. Wallach, R. Fergus, S.
  Vishwanathan, and R. Garnett, editors, {\em Advances in Neural Information
  Processing Systems}, volume~30. Curran Associates, Inc., 2017.

\bibitem{snell2017prototypical}
J. Snell, K. Swersky, and R. Zemel.
\newblock Prototypical networks for few-shot learning.
\newblock In I. Guyon, U.~V. Luxburg, S. Bengio, H. Wallach, R. Fergus, S.
  Vishwanathan, and R. Garnett, editors, {\em Advances in Neural Information
  Processing Systems}, volume~30. Curran Associates, Inc., 2017.

\bibitem{Spinoulas2020}
Leonidas Spinoulas, Mohamed~E. Hussein, David Geissbühler, Joe Mathai,
  Oswin~G. Almeida, Guillaume Clivaz, Sébastien Marcel, and Wael~Abd Almageed.
\newblock Multispectral biometrics system framework: Application to
  presentation attack detection.
\newblock {\em IEEE Sensors Journal}, pages 1--1, 2021.

\bibitem{stan2021unsupervised}
Serban Stan and Mohammad Rostami.
\newblock Unsupervised model adaptation for continual semantic segmentation.
\newblock In {\em Proceedings of the AAAI Conference on Artificial
  Intelligence}, volume~35, pages 2593--2601, 2021.

\bibitem{swir_face_pad}
Holger Steiner, Sebastian Sporrer, Andreas Kolb, and Norbert Jung.
\newblock Design of an {Active} {Multispectral} {SWIR} {Camera} {System} for
  {Skin} {Detection} and {Face} {Verification}.
\newblock {\em Journal of Sensors}, 2016:16, 2016.

\bibitem{wang2020deep}
Z. {Wang}, Z. {Yu}, C. {Zhao}, X. {Zhu}, Y. {Qin}, Q. {Zhou}, F. {Zhou}, and Z.
  {Lei}.
\newblock Deep spatial gradient and temporal depth learning for face
  anti-spoofing.
\newblock In {\em 2020 IEEE/CVF Conference on Computer Vision and Pattern
  Recognition (CVPR)}, pages 5041--5050, 2020.

\bibitem{xie2019attentive}
G. {Xie}, L. {Liu}, X. {Jin}, F. {Zhu}, Z. {Zhang}, J. {Qin}, Y. {Yao}, and L.
  {Shao}.
\newblock Attentive region embedding network for zero-shot learning.
\newblock In {\em 2019 IEEE/CVF Conference on Computer Vision and Pattern
  Recognition (CVPR)}, pages 9376--9385, 2019.

\bibitem{xie2020region}
G.-S. Xie, L. Liu, F. Zhu, F. Zhao, Z. Zhang, Y. Yao, J. Qin, and L. Shao.
\newblock Region graph embedding network for zero-shot learning.
\newblock In A. Vedaldi, H. Bischof, T. Brox, and J.-M. Frahm, editors, {\em
  Computer Vision -- ECCV 2020}, pages 562--580, Cham, 2020. Springer
  International Publishing.

\bibitem{yang2019face}
X. {Yang}, W. {Luo}, L. {Bao}, Y. {Gao}, D. {Gong}, S. {Zheng}, Z. {Li}, and W.
  {Liu}.
\newblock Face anti-spoofing: Model matters, so does data.
\newblock In {\em 2019 IEEE/CVF Conference on Computer Vision and Pattern
  Recognition (CVPR)}, pages 3502--3511, 2019.

\bibitem{yu2019unsupervised}
Q. {Yu} and K. {Aizawa}.
\newblock Unsupervised out-of-distribution detection by maximum classifier
  discrepancy.
\newblock In {\em 2019 IEEE/CVF International Conference on Computer Vision
  (ICCV)}, pages 9517--9525, 2019.

\bibitem{zitong2020}
Z. Yu, X. Li, X. Niu, J. Shi, and G. Zhao.
\newblock Face anti-spoofing with human material perception.
\newblock In A. Vedaldi, H. Bischof, T. Brox, and J.-M. Frahm, editors, {\em
  Computer Vision -- ECCV 2020}, pages 557--575, Cham, 2020. Springer
  International Publishing.

\bibitem{yu2020searching}
Z. {Yu}, C. {Zhao}, Z. {Wang}, Y. {Qin}, Z. {Su}, X. {Li}, F. {Zhou}, and G.
  {Zhao}.
\newblock Searching central difference convolutional networks for face
  anti-spoofing.
\newblock In {\em 2020 IEEE/CVF Conference on Computer Vision and Pattern
  Recognition (CVPR)}, pages 5294--5304, 2020.

\bibitem{casia_surf}
S. {Zhang}, A. {Liu}, J. {Wan}, Y. {Liang}, G. {Guo}, S. {Escalera}, H.~J.
  {Escalante}, and S.~Z. {Li}.
\newblock {CASIA-SURF: A Large-Scale Multi-Modal Benchmark for Face
  Anti-Spoofing}.
\newblock {\em IEEE Transactions on Biometrics, Behavior, and Identity
  Science}, 2(2):182--193, 2020.

\bibitem{Zhang2000}
Z. {Zhang}.
\newblock A flexible new technique for camera calibration.
\newblock {\em IEEE Transactions on Pattern Analysis and Machine Intelligence},
  22(11):1330--1334, Nov 2000.

\end{thebibliography}
}

\clearpage
\begin{appendices}
 \section{PADISI-Face Dataset}

Previous efforts on generating face PAD datasets have been focused on a number of major attack types, including, disguises, printed photographs, 3D masks, and replays. As novel types of  attacks emerge, existing datasets might be insufficient to guarantee designing suitable PAD algorithms because there has not been enough effort in generating datasets that cover a wide variety of attack categories. Table~\ref{tab:table1} summarizes a number of highly prevalent face PAD datasets in the literature. The table provides information about the types of presentation attack instruments (PAIs) present in each dataset. As it can be seen, these datasets are limited in terms of diversity of attack types they contain. PADISI-Face is a new dataset  captured from $182$ different participants to offer more diverse set of attack types. Due to granular labels on these attack types, PADISI-Face  is a suitable dataset for testing models in continual learning settings or when there should be  significant variations between the training and testing datasets.

\begin{figure} 
    \centering
    \includegraphics{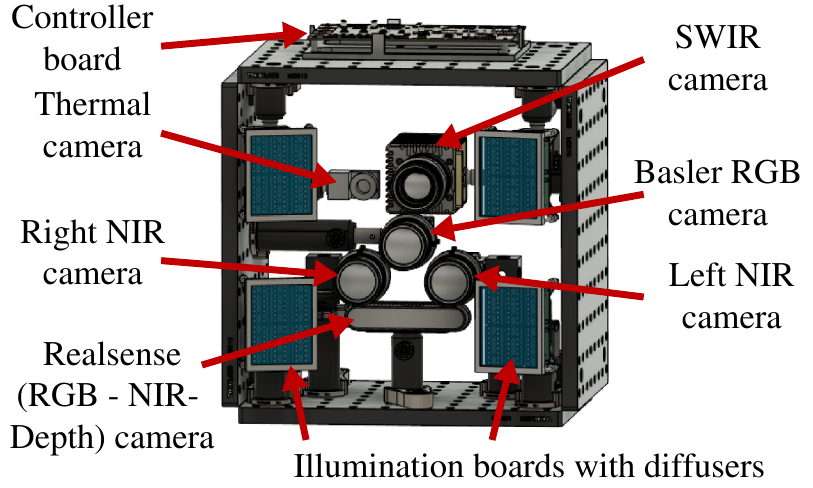}
    \caption{Face biometric sensor suite.}
    \label{fig:system}
\end{figure}

\begin{figure*}
\includegraphics{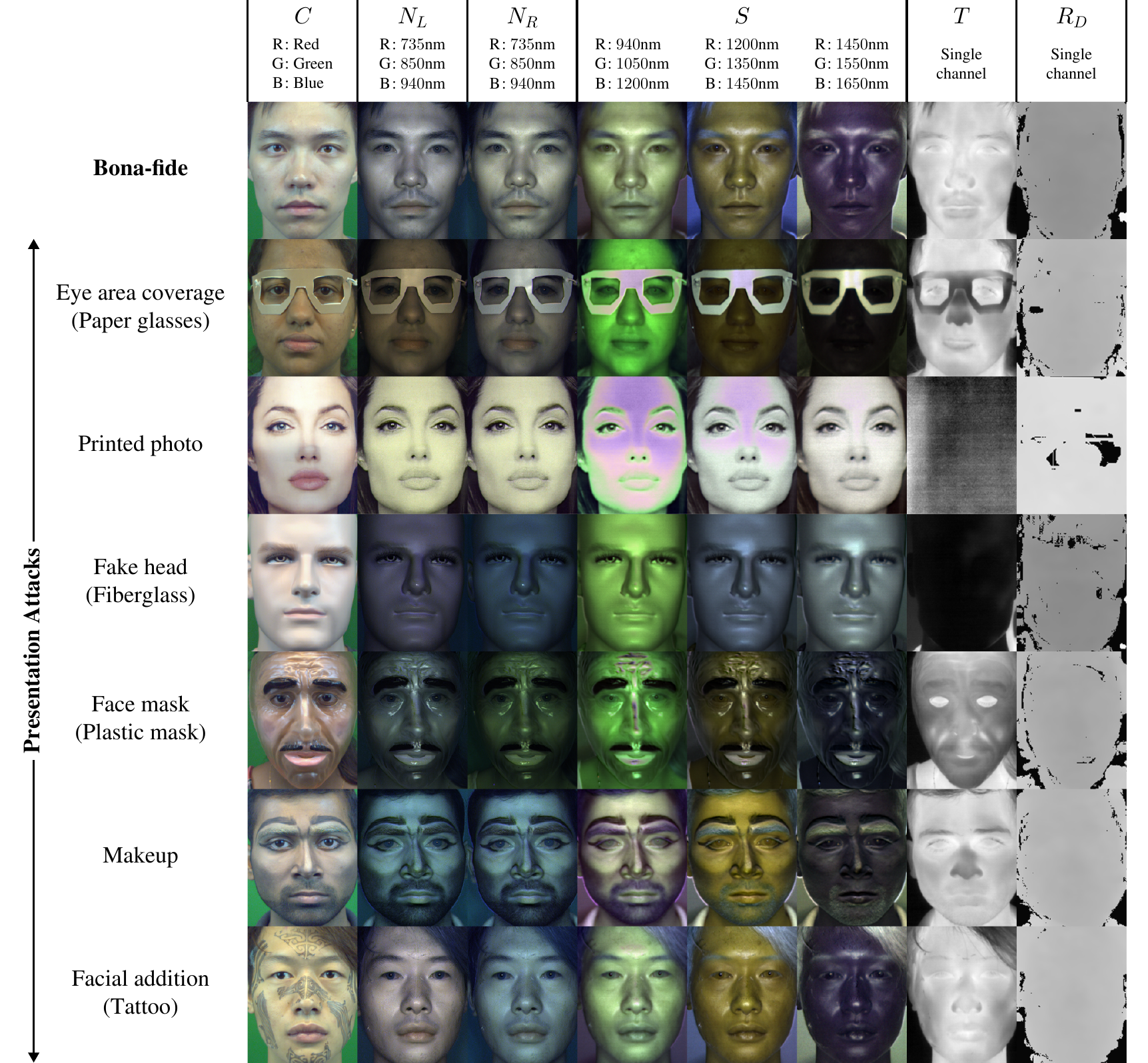}
\caption{RGB visualization of samples collected for all types of PAIs are summarized in this figure. For each image, the corresponding dark channel has been subtracted and each RGB channel has been min-max normalized for visualization purposes. Note, not all images have the same resolution but are resized for fine arrangement.}
\label{fig:multispectral_images}
\end{figure*}

The PADISI-Face Dataset is collected using a sensor array designed and built by our team, shown in Figure~\ref{fig:system}~\cite{Spinoulas2020}. The system is designed for more comprehensive future versions of PADISI-Face  that will contain beyond the visible range information. The hardware  comprises of six different cameras spanning visible (RGB), short-wave-infrared (SWIR) and long-wave infrared (Thermal) electromagnetic spectrum ranges. Additionally, there are two near-infrared (NIR) cameras for high quality stereo depth estimation. For acquisition of data in NIR and SWIR spectra, a synchronized illumination of different wavelength LEDs (shown in Figure~\ref{fig:system}) were used. The synchronized sequence of LED illuminations were designed to maximize the throughput of the camera suite while increasing the temporal coherence between frames. Figure~\ref{fig:multispectral_images} shows some examples of images collected for bona-fide and several attack samples using the sensor array in various light ranges. For NIR and SWIR modalities, dark channel subtraction is performed to reduce the effect of ambient illumination. Data was collected from each participant over two rounds. In the first round, bona fide samples were collected. Participants presented a presentation attack instrument (PAI) in the second round. PADISI-Face will be available for the use of the research community.

\begin{table} 
\scriptsize
  \begin{center}
    \caption{Multi-spectral PAD Datasets}
    \label{tab:table1}
    \begin{tabular}{c||c|c|c}
    \toprule \hline
      \textbf{Dataset} & \textbf{Year} & \textbf{Participants} & \textbf{Attacks} \\ \hline \hline
      Pavlidis Symosek~\cite{nir_dataset_01} & $2000$ & $-$ & \multirow{1}{*}{Facial disguises}   \\
      \hline
      3DMAD~\cite{3dmad} & $2013$ & $17$ & 3D mask attacks \\
      \hline
      ${I^2}$BVSD~\cite{I2BVSD} & $2013$ & $75$ & Facial disguises \\
      \hline
      GUC-LiFFAD~\cite{GUC-LiFFAD} & $2015$ & $80$ & 2D print and replay \\
      \hline
      MS-Spoof~\cite{msspoof-2015} & $2015$ & $21$ & 2D print  \\
      \hline
      BRSU~\cite{swir_face_pad} & $2016$ & $50$ & 3D masks \\
      \hline
       EMSPAD~\cite{emspad} & $2017$ & $50$ & 2D print  \\
      \hline
      MLFP~\cite{mlfp} & $2017$ & $10$ & 2D \& 3D masks  \\
      \hline
      CASIA-SURF~\cite{casia_surf} & $2020$ & $1000$ & 2D print \& cutouts \\
      \hline
      \multirow{4}{*}{\textbf{\ds}} & \multirow{4}{*}{$2020$} & \multirow{4}{*}{$360$} & 2D print, mannequins   \\
      & & & 3D masks,    \\
      & & & obfuscation makeup 
      \\
      & & & fake tattoo, eye area cover   
      \\ 
      \hline \bottomrule
    \end{tabular}
  \end{center}
\end{table}

To enable face detection in all captured frames, we use a standard calibration process using checkerboards~\cite{Zhang2000}. For the checkerboard to be visible in all wavelength regimes, a manual approach is used when a sequence of frames is captured offline while the checkerboard is being lit with a bright halogen light. This makes the checkerboard pattern visible and detectable by all cameras which allows the standard calibration estimation process to be followed. The face can then be easily detected in the RGB space~\cite{fan} and the calculated transformation for each camera can be applied to detect the face in the remaining camera frames.

Following the aforementioned approach, face landmarks are detected on the visible spectrum using~\cite{fan}. A bounding box is then constructed from the landmarks to have a tight crop of the face. The bounding box of the visible spectrum is then projected to the corresponding co-ordinate system of the other cameras to extract approximately aligned faces on different modalities. Each channel is then scaled to the range $[0, 1]$ by dividing the bit depth of the camera and then resized to $160\times 160$ pixels. In our experiments, we use the visible range information as input to our algorithm. A future direction includes considering beyond the visible information to perform PAD.

\section{Reducing False-Positive Predictions}
 
 Our primary focus has been on reducing the false-negative predictions.  In practical settings, we can assume labels for detected novel data points can be accessible with a delay by the end of each task, e.g., using manual annotation. To model this possibility, we performed an experiment using a selection scheme that assumes manual annotation is possible, i.e., label pollution is reduced to zero. Since updating the model occurs at discrete periods at the end of each task, we have assumed the delay for labeling is less than the time needed to update the model.  Hence by the time a task finishes, we assume the labels for the novel samples are accessible before updating the model. Table~\ref{tab:tab5} presents  results for the PADISI-Face dataset in the single PA/task scenario, where we have compared Delayed Labels (DL) with NACL.   We observe that this   sampling scheme leads to reduced false-negative predictions. Additionally, we observe BPCER performance also improves. 
  \begin{table}[!t]
   \caption{Effect of knowing labels for the challenging data points using the   PADISI-Face dataset in the single PA/task scenario. Results for NACL are taken from the bottom left of Fig. 5.}
    \vspace{2mm}
    \label{tab:continualbatlabl}
    \centering
    \begin{scriptsize}
    \begin{tabular}{c|cc|cc|cc} \hline
        
         Task     &  \multicolumn{2}{c|}{APCER ($\%$)}     & \multicolumn{2}{c|}{BPCER ($\%$)}      & \multicolumn{2}{c}{ACER ($\%$)}\\
         \hline 
         No.      &  NACL    & DL   &  NACL    & DL&  NACL    & DL  \\   \hline
$ 1 $
& $ 46.1 $ & $ 27.8 $ & $ 10.7 $ & $ 4.8 $ & $ 28.4 $ & $ 16.3 $
\\
$ 2 $
& $ 38.2 $ & $ 24.8 $ & $ 10.1 $ & $ 1.0 $ & $ 24.1 $ & $ 12.9 $
\\
$ 3 $
& $ 41.4 $ & $ 27.0 $ & $ 14.1 $ & $ 1.5 $ & $ 27.8 $ & $ 14.2 $
\\
$ 4 $
& $ 29.1 $ & $ 19.9 $ & $ 15.0 $ & $ 1.1 $ & $ 22.0 $ & $ 10.5 $
\\
$ 5 $
& $ 15.1 $ & $ 12.5 $ & $ 18.1 $ & $ 4.3 $ & $ 16.6 $ & $ 8.4 $
\\
$ 6 $
& $ 19.4 $ & $ 16.0 $ & $ 18.0 $ & $ 3.7 $ & $ 18.7 $ & $ 9.8 $
\\
$ 7 $
& $ 17.2 $ & $ 15.1 $ & $ 17.0 $ & $ 2.6 $ & $ 17.1 $ & $ 8.9 $
\\
$ 8 $
& $ 18.2 $ & $ 13.8 $ & $ 17.4 $ & $ 2.8 $ & $ 17.8 $ & $ 8.3 $
\\
 \hline  
    \end{tabular}
    \end{scriptsize}
    \label{tab:tab5}
\end{table}

 \section{Experimental Setup Details}
 
 We provide details that we used to perform   experiments.

 \subsection{Network structure}
 In our experiments, the network is consisted of a pre-trained fixed backbone encoder, followed by fully connected layers to reach to the label space.

 \textbf{Backbone Model} 
 
\begin{figure}[!t]
    \centering
    \includegraphics[scale=0.5]{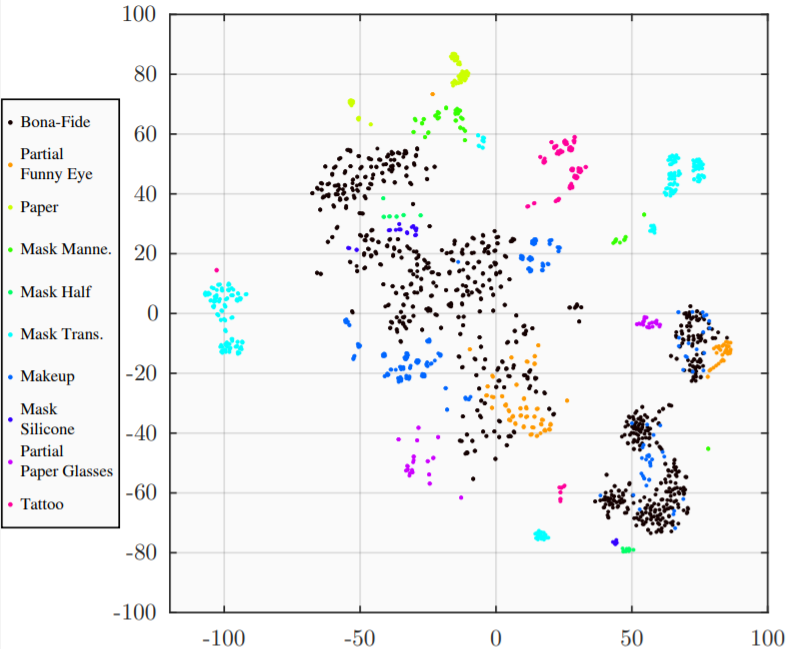}
    \caption{t-SNE visualization of features obtained using the pre-trained model of~\cite{he2020momentum} (MoCo-v1), for PADISI-Face dataset. One frame, per capture, is used for this visualization.}
    \vspace{-0.1cm}
    \label{fig:tsne_features}
\end{figure}

 An important limitation of CNN models when trained on small datasets, such as biometric datasets, is that they tend to select features which are not generalizable due to overfitting.
For this purpose, we opted for employing MoCo-v1 as a fixed backbone network~\cite{he2020momentum} to improve generalizability of the extracted features. This network is trained on ImageNet using a contrastive loss that attempts to find similarities and dissimilarities among synthesized variants of training data samples in an unsupervised way. It is subclass of self-supervised learning at which a deep neural network is trained to solve pseudo-tasks. 
As a result, the network learns to extract discriminative features at its early layers to solve the pseudo-tasks. Thus, when we use MoCo-v1 as our backbone for PAD in a continual learning setting, no input or label information of the future PA types have been used to train the feature extraction model. This property ensures that no information about the training dataset has been used in training the model. This allows to claim that the new attacks are indeed unseen. We note that extracting features using this pre-trained network leads to an separability of different types of attacks, as shown in the t-SNE visualizations of Figure~\ref{fig:tsne_features}, for PADISI-Face   dataset as an example which enables our model to identify data points that belong to new attack classes. This observation demonstrates we can use the backbone model as a good feature extractor to identify OTDS.

 \textbf{Learnable Layers} 
 
 We use the same  end-to-end network structure  for a fair comparison among the methods. The MoCo  backbone is followed by three fully connected layers with $64$, $32$, and $2$ (turn into $3$ nodes when the model is trained to identify OTDS) nodes each. MoCo's encoder is in essence a ResNet50 architecture with $128$ output nodes, used here as discriminative feature vectors to improve classification. In all experiments, the weights of the backbone network are frozen and learnable parameters $\theta$ in our formulated would refer to the last fully connected layers. We use ReLU non-linearity in the first two layers and softmax non-linearity in the final layer. We have selected the layer with $32$ nodes to represent the embedding space $\mathcal{Z}$ on which the CL approach is performed, as described in the paper. At each task, the network is trained with 2 output nodes and performance on the testing split is measured during training. When the task is learned, the network output is extended to include a third output. After identifying the OTDS, the network again is trained with 2 outputs. This process is continued until all tasks are learned. To reduce redundancy of inference at stochastic gradient descent step, we compute the input features initially and perform optimization just on the learnable layers. By doing so, we reduce learning time but have the understanding that in practice, inference also needs to be performed end-to-end.

\subsection{Implementation Parameters} 

We use the cross entropy loss as the discrimination loss.  We used Keras for implementation of the algorithm and  the Adam optimizer to perform stochastic gradient descent. The learning rate is set to be $2\times10^{-4}$ with a decay rate of $10^{-4}$. We use a batch size of $100$.
At each batch, we select 100 points randomly and make sure the batch is balanced. To learn each task, we randomly initialize all the trainable weights (fully connected layers) and perform optimization using 10000 batches.
At each   training epoch, we computed the loss function on the training data split and the performance metrics on the testing split. We ran our code on a cluster node equipped with 4 Nvidia Tesla P100-SXM2 GPU's.  Our code is provided as part of the supplementary material.
  
\end{appendices}
  
\end{document}